# UCP-Networks: A Directed Graphical Representation of Conditional Utilities

**Craig Boutilier**
Department of Computer Science
University of Toronto
Toronto, ON M5S 3H5
cebly@cs.toronto.edu

**Fahiem Bacchus**
Department of Computer Science
University of Toronto
Toronto, ON M5S 3H5
fbacchus@cs.toronto.edu

**Ronen I. Brafman**
Department of Computer Science
Ben-Gurion University
Beer Sheva, Israel 84105
brafman@cs.bgu.ac.il

## Abstract

We propose a directed graphical representation of utility functions, called UCP-networks, that combines aspects of two existing preference models: generalized additive models and CP-networks. The network decomposes a utility function into a number of additive factors, with the directionality of the arcs reflecting conditional dependence in the underlying (qualitative) preference ordering under a *ceteris paribus* interpretation. The CP-semantics ensures that computing optimization and dominance queries is very efficient. We also demonstrate the value of this representation in decision making. Finally, we describe an interactive elicitation procedure that takes advantage of the linear nature of the constraints on "tradeoff weights" imposed by a UCP-network.

## 1 Introduction

Effective representations for preferences and utility functions are critical to the success of many AI applications. A good preference or utility representation should capture statements that are natural for users to assess, or are easy to learn from data; it should offer the compact expression of preferences or utilities; and it should support effective inference.

A useful design stance for such representation is to exploit the structure of utility functions using notions from multiattribute utility theory, such as conditional preferential independence, mutual utility independence, etc. [10]. Recent work has exploited such structure to develop graphical models: Bacchus and Grove [1, 2] propose an undirected network representation for (quantitative) utility that captures conditional additive utility independencies; Boutilier, Brafman, Hoos and Poole [3] propose a directed network representation for (qualitative) preference functions that captures conditional preference statements under a *ceteris paribus* (all else equal) assumption. La Mura and Shoham [11] describe a hybrid representation for combining both probabilistic and utility information in a undirected graphical model representing *expected* utilities directly.

In this paper, we propose a new directed network representation for utility functions that combines certain aspects of the first two of these approaches. The *UCP-network* formalism can be viewed as an extension of the CP-network model [3] that allows one to represent quantitative utility information rather than simple preference orderings. The formalism also utilizes the notion of generalized additive independence (GAI) [1]. By employing a directed graph, UCP-nets allow one to make more powerful statements that are often more natural and lead to more effective inferences. In particular, we will show that dominance and optimization queries can be answered directly in UCP-nets. In addition, the formalism can be used in an interactive elicitation process to determine *relevant* parameters of the UCP model in a specific decision scenario. We propose a technique for elicitation—much like that proposed by Chajewska, Koller, and Parr [4]—that exploits the linear constraints imposed by a partially-specified UCP-model to determine an "optimal" sequence of queries.

The rest of this paper is organized as follows: Section 2 provides necessary background. Section 3 describes UCP-nets, their properties, and their relation to GAI decompositions of utility functions and CP-networks. Section 4 discusses the problem of optimization in the context of UCP-nets, and shows the advantage of this representation tool. Section 5 explains how elicitation and optimization can be performed concurrently in order to recognize near-optimal choices with minimal questioning. We conclude in Section 6 with a discussion of future work.

## 2 Background Concepts

We begin with an outline of some relevant notions from multiattribute utility theory [10]. We assume a set of actions $\mathcal{A}$ is available to a decision maker, each action having one of a number of possible *outcomes*. The set of all outcomes is designated $\mathcal{O}$. A *preference ranking* is a total preorder $\succeq$ over the set of outcomes: $o_1 \succeq o_2$ means that outcome $o_1$ is equally or more preferred by the decision maker than $o_2$. A *utility function* is a bounded, real-valued function $u : \mathcal{O} \mapsto \mathbf{R}$. A utility function $u$ induces a preference ordering $\succeq$ such that $o_1 \succeq o_2$ iff $u(o_1) \geq u(o_2)$. A utility function also induces preferences over *lotteries*, or distributions over outcomes, where one lottery is preferred to another when its expected utility is greater. When actions have uncertain outcomes, thereby generating a distribution over outcomes, preferences for actions can be equated with preferences for the corresponding lotteries [12].

One difficulty encountered in eliciting, representing, and reasoning with preferences and utilities is the size of the outcome space, which is generally determined by a set of variables. We assume a set of variables $\mathbf{V} = \{X_1, \ldots, X_n\}$ characterizing possible outcomes. Each variable $X_i$ has domain $Dom(X_i) = \{x_1^i, \ldots, x_{n_i}^i\}$. The set of outcomes is $\mathcal{O} = Dom(\mathbf{V}) = Dom(X_1) \times \cdots \times Dom(X_n)$. Thus direct assessment of a preference function is generally infeasible due to the exponential size of $\mathcal{O}$. Fortunately, a preference function can be specified concisely if it exhibits sufficient structure. We describe certain standard types of structure here (see [10] for further details).

We denote a particular assignment of values to a set $\mathbf{X} \subseteq \mathbf{V}$ as $\mathbf{x}$, and the concatenation of two non-intersecting partial assignments by $\mathbf{xy}$. If $\mathbf{X} \cup \mathbf{Y} = \mathbf{V}$, $\mathbf{xy}$ is a complete outcome, and $\mathbf{xy}$ is a *completion* of the partial assignment $\mathbf{x}$. $Comp(\mathbf{x})$ denotes the set of completions of $\mathbf{x}$.

A set of features $\mathbf{X}$ is *preferentially independent* of its complement $\mathbf{Y} = \mathbf{V} - \mathbf{X}$ iff, for all $\mathbf{x}_1, \mathbf{x}_2, \mathbf{y}_1, \mathbf{y}_2$, we have

$$\mathbf{x}_1\mathbf{y}_1 \succeq \mathbf{x}_2\mathbf{y}_1 \text{ iff } \mathbf{x}_1\mathbf{y}_2 \succeq \mathbf{x}_2\mathbf{y}_2$$

We denote this as $PI(\mathbf{X}, \mathbf{Y})$. In other words, the structure of the preference relation over assignments to $\mathbf{X}$, when all other features are held fixed, is the same no matter what values these other features take. If $PI(\mathbf{X}, \mathbf{Y})$ and $\mathbf{x}_1\mathbf{y} \succeq \mathbf{x}_2\mathbf{y}$ for any assignment $\mathbf{y}$ to $\mathbf{V} - \mathbf{X}$, then we say that $\mathbf{x}_1$ is preferred to $\mathbf{x}_2$ *ceteris paribus*. Thus, one can assess the relative preferences over assignments to $\mathbf{X}$ once, knowing these preferences do not change as other attributes vary. We define conditional preferential independence analogously. Let $\mathbf{X}$, $\mathbf{Y}$, and $\mathbf{Z}$ be nonempty sets that partition $\mathbf{V}$. $\mathbf{X}$ and $\mathbf{Y}$ are *conditionally preferentially independent* given an assignment $\mathbf{z}$ to $\mathbf{Z}$ (denoted $CPI(\mathbf{X}, \mathbf{z}, \mathbf{Y})$) iff, for all $\mathbf{x}_1, \mathbf{x}_2, \mathbf{y}_1, \mathbf{y}_2$, we have

$$\mathbf{x}_1\mathbf{z}\mathbf{y}_1 \succeq \mathbf{x}_2\mathbf{z}\mathbf{y}_1 \text{ iff } \mathbf{x}_1\mathbf{z}\mathbf{y}_2 \succeq \mathbf{x}_2\mathbf{z}\mathbf{y}_2$$

In other words, the preferential independence of $\mathbf{X}$ and $\mathbf{Y}$ holds when $\mathbf{Z}$ is assigned $\mathbf{z}$. If we have $CPI(\mathbf{X}, \mathbf{z}, \mathbf{Y})$ for all $\mathbf{z} \in Dom(\mathbf{Z})$, then $\mathbf{X}$ and $\mathbf{Y}$ are *conditionally preferentially independent* given $\mathbf{Z}$, denoted $CPI(\mathbf{X}, \mathbf{Z}, \mathbf{Y})$.

Decomposability of a preference function often allows one to identify the most preferred outcomes rather readily. Unfortunately, the *ceteris paribus* component of these definitions means that the *CPI* statements are relatively weak. In particular, they do not imply a stance on specific value tradeoffs. For instance, suppose $PI(A, B)$ and $PI(B, A)$ so that the preferences for values of $A$ and $B$ can be assessed separately, with $a_1 \succ a_2$ and $b_1 \succ b_2$. Clearly, $a_1b_1$ is the most preferred outcome and $a_2b_2$ is the least; but if feasibility constraints make $a_1b_1$ impossible, we must be satisfied with one of $a_1b_2$ or $a_2b_1$. With just preferential independence we cannot tell which is most preferred using these separate assessments. Stronger conditions (e.g.,

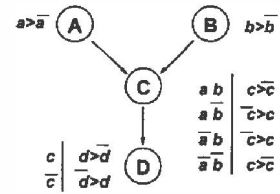

Figure 1: A CP-Network

*mutual preferential independence* [10]) are required before such tradeoffs can be easily evaluated.

CP-nets [3] are a graphical representation for structuring *CPI* statements. In particular, CP-nets are directed acyclic graphs whose nodes are the variables of $\mathbf{V}$. We associate a *conditional preference table* (CPT) with each node $X$ specifying a preference order over $X$'s values given each instantiation of its parents $\mathbf{U}$, and require that $CPI(X, \mathbf{U}, \mathbf{Z})$ hold, where $\mathbf{Z} = \mathbf{V} - (\mathbf{U} \cup \{X\})$. CP-nets structure these *CPI* statements so as to support useful inferences about the underlying preference order [3]. In Fig. 1 we see a CP-net defined over a set of four boolean variables, where, e.g., the CPT for $C$ specifies that $c$ is preferred to $\bar{c}$ when $a$ and $b$ hold. An important property of CP-nets is the induced importance it assigns to different variables: nodes "higher-up" in the graph are more important than their descendants. Thus, it is more important to obtain preferred values for a node than for any one of its descendants. For example, in the CP-net above, we can see that $ab\bar{c}\bar{d}$ (in which a less preferred value of $C$ appears) is preferred to $\bar{a}b\bar{c}\bar{d}$ (in which a less preferred value of $A$ appears). This property plays an important role in UCP-nets.

Let $\mathbf{X}_1, \ldots, \mathbf{X}_k$ be sets of *not necessarily disjoint* variables such that $\mathbf{V} = \cup_i \mathbf{X}_i$. $\mathbf{X}_1, \ldots, \mathbf{X}_k$ are *generalized additive independent (GAI)* for an underlying utility function $u$ if, for any two probability distributions $Pr_1$ and $Pr_2$ over $Dom(\mathbf{V})$ that have the same marginals on each of the sets of variables $\mathbf{X}_i$, $u$ has the same expected value under $Pr_1$ and $Pr_2$. In other words, the expected value of $u$ is not affected by correlations between the $\mathbf{X}_i$. It depends only on the the marginal distributions over each the $\mathbf{X}_i$.

It can be shown [1] that $\mathbf{X}_1, \ldots, \mathbf{X}_k$ are GAI *iff* $u$ can be written as $u(\mathbf{V}) = \sum_{i=1}^{k} f_i(\mathbf{X}_i)$. That is, $u$ can be decomposed into a sum of factors over each of these sets of variables. This property generalizes the standard definition of additive utility independence, which requires that the $\mathbf{X}_i$ partition $\mathbf{V}$. For UCP-nets the ability to deal with overlapping sets of variables is critical.

## 3 Adding Utilities to CP-Nets

As noted above, the precision of a utility function (as opposed to a preference ordering) is often needed in decision making contexts where uncertainty is a factor. The representation of utility functions should be natural, easy to elicit, compact (in typical cases), and support effective inference. There are two basic types of queries with regard

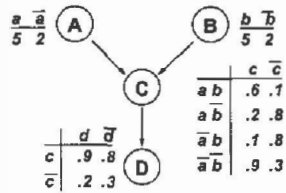

Figure 2: A UCP-Network

to outcomes one will often ask:[1]

(a) *Dominance queries*: does one outcome have higher utility than another (i.e., $u(\mathbf{v}_1) \geq u(\mathbf{v}_2)$)?

(b) *Outcome optimization queries*: what outcome has maximum utility given some partial assignment (i.e., what is $\arg\max\{u(\mathbf{v}) : \mathbf{v} \in Comp(\mathbf{x})\}$)?

GAI-models allow dominance testing to be performed very effectively: the utility of outcome $\mathbf{v}$ is readily determined by looking up the value $f_i(\mathbf{v})$ of each factor applied to $\mathbf{v}$, and summing them to obtain $u(\mathbf{v})$. In contrast, CP-nets do not allow straightforward dominance tests, generally requiring reasonably sophisticated search techniques for all but the simplest network topologies. The relative attractiveness of the two approaches is, however, reversed when one considers optimization queries. In CP-nets, determining the (conditional) maximal outcome in a preference relation is straightforward. In contrast, maximization in a GAI model requires the use of variable elimination [6, 13], whose complexity depends on the structure of the model.[2]

We propose in this section a new network representation for utilities that combines aspects of both CP-nets and GAI models. The model is directed, like CP-nets, but preferences are quantified with utilities. The semantics is given by generalized additive independence along with the constraint that the directed influences reflect the *ceteris paribus* condition underlying CP-nets. By extending CP-nets with quantitative utility information, expressive power is enhanced and dominance queries become computationally efficient. By introducing directionality and a *ceteris paribus* semantics to the GAI model, we allow utility functions to be expressed more naturally, and permit optimization queries to be answered much more effectively.

A UCP-net extends a CP-net by allowing quantification of nodes with conditional utility information. Semantically, we treat the different factors as generalized additive independent of one another. For example, the network in Fig. 1 can be extended with utility information by including a factor for each family in the network, specifically, $f_1(A)$, $f_2(B)$, $f_3(A, B, C)$, and $f_4(C, D)$ (see Fig. 2). We interpret this network using GAI: $u(A, B, C, D) = f_1(A) + f_2(B) + f_3(A, B, C) + f_4(C, D)$. Each of these factors is quantified by the (now quantitative) CPT tables in the network. For example, in Fig. 2 we have that $f_3(a, b, c) = 0.6$, while $f_3(a, b, \bar{c}) = 0.1$. Thus, the CPT tables along with the GAI interpretation provide a full specification of the utility function. For example, we have that $u(a, b, \bar{c}, \bar{d}) = f_1(a) + f_2(b) + f_3(a, b, \bar{c}) + f_4(\bar{c}, \bar{d}) = 5 + 5 + 0.1 + 0.3 = 10.4$.

---
[1] Queries regarding optimization with respect to actions are discussed in the next section.
[2] GAI optimization can be effected using *cost networks* [7].

Notice however that the factors $f_1$ and $f_2$ are redundant in the sense that they refer to variables that are included in $f_3$. Thus, this utility function could be represented more concisely using a GAI decomposition containing two factors: $f_4(C, D)$ and $f_5(A, B, C) = f_1(A) + f_2(B) + f_3(A, B, C)$. The directionality of the utility-augmented CP-network, on the surface, seems to serve no purpose other than to break up the GAI-factors unnecessarily.

However, we can use this directionality to represent CP conditions on the utility function $u$, and thus provide a simple and natural interpretation for the individual factors of $u$. In particular, we interpret the fact that $A$ and $B$ are parents of $C$ as asserting that $CPI(C, \{A, B\}, D)$, and thus the factor $f_3(A, B, C)$ specifies the utility of $C$ *given* $A$ and $B$. The fact that each node is isolated from the rest of the network given the values of its parents greatly simplifies utility assessment. Furthermore, this structure supports more efficient inference for certain queries than the standard GAI representation.

**Definition 1** Let $u(X_1, \ldots, X_n)$ be a utility function with induced preference relation $\succeq$. A *UCP-network for $u$* is a DAG $G$ over $X_1, \ldots, X_n$ and a quantification (i.e., a set of factors $f_i(X_i, \mathbf{U}_i)$ where $\mathbf{U}_i$ are the parents of $X_i$) s.t.:

(a) $u(X_1, \ldots, X_n) = \sum_i f_i(X_i, \mathbf{U}_i)$

(b) The DAG $G$ is a valid CP-network for $\succeq$; i.e., $\succeq$ satisfies $CPI(X_i, \mathbf{U}_i, \mathbf{Z}_i)$ for each $X_i$, where $\mathbf{Z}_i = \mathbf{V} - (\mathbf{U}_i \cup \{X_i\})$

Condition (a) means that every UCP-net specifies a GAI decomposition of the underlying utility function $u$. However, the acyclic restriction means that not every GAI decomposition can be represented in a UCP-net. For example, the GAI decomposition $u(A, B, C) = f_1(A, B, C) + f_2(C, B)$ would have to converted to the decomposition $u(A, B, C) = f_3(A, B, C)$, where $f_3 = f_1 + f_2$, before it could be represented as a UCP-net. Nevertheless, there is a simple case where a GAI decomposition can easily be seen to be representable using a UCP-net topology.

**Proposition 1** *If there exists a ordering of the variables such that under this ordering the last variable in every GAI factor is unique (i.e., no variable is last in more than one factor), then the GAI decomposition can be represented with a UCP-net topology.*

To construct the UCP-net in this case, for every factor we make every "last" variable a child and all of the prior variables its parents.

Even if we can represent a GAI decomposition in a UCP-net topology and we parameterize the net using the GAI

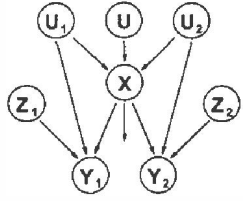

Figure 3: The Domination Relation

factors, the the result might not be a UCP-net, since the utility function might not satisfy the CP requirements of condition (b). For example, let $u$ be a utility function over the boolean variables $A$ and $B$ with $u(ab) = 9, u(a\bar{b}) = 1, u(\bar{a}b) = 2$ and $u(\bar{a}\bar{b}) = 8$. No UCP-net can represent $u$. To see this, first notice $PI(A, B)$ fails to holds, since preferences for $B$ depend on $A$. But we cannot make $A$ a parent of $B$, since the preferences for $A$ depend on $B$; nor can we make $B$ a parent of $A$ for a similar reason.[3]

This example shows that, for a fixed set of variables, UCP-nets define a proper subset of all utility functions; but this subset has certain attractive computational properties as well as pragmatic advantages when it comes to elicitation.

Given a UCP-network $\mathcal{U}$ for utility function $u$, verifying that it satisfies the CP-relationships among variables required by its definition can be accomplished by tests involving the local neighborhoods of each node in the network.

**Definition 2** Let $X$ be a variable in a quantified DAG with parents $\mathbf{U}$ and children $\mathbf{Y} = \{Y_1, \ldots, Y_n\}$, and let $\mathbf{Z}_i$ be the parents of $Y_i$, excluding $X$ and any elements of $\mathbf{U}$. Let $\mathbf{Z} = \cup \mathbf{Z}_i$. Let $\mathbf{U}_i$ be the subset of variables in $\mathbf{U}$ that are parents of $Y_i$ (the relationships among these variables is shown in Fig. 3). We say $X$ *dominates its children given* $\mathbf{u} \in Dom(\mathbf{U})$ if, for all $x_1, x_2$ such that $f_X(x_1, \mathbf{u}) \geq f_X(x_2, \mathbf{u})$, for all $\mathbf{z} \in Dom(\mathbf{Z})$, and for all $\langle y_1, \ldots, y_n \rangle \in Dom(\mathbf{Y})$:

$$f_X(x_1, \mathbf{u}) - f_X(x_2, \mathbf{u}) \geq \sum_i f_{Y_i}(y_i, x_2, \mathbf{u}_i, \mathbf{z}_i) - f_{Y_i}(y_i, x_1, \mathbf{u}_i, \mathbf{z}_i)$$

$X$ *dominates its children* if this holds for all $\mathbf{u} \in Dom(\mathbf{U})$.

Testing whether $X$ dominates its children is a local test, involving only the factor for variable $X$ and those of its children. It requires that we check that, for each instantiation of $X$'s children and the parents of its children, whether the decrease in *local utility* (i.e., in factor $f_X$) dominates the (potential) increase it causes in the local utilities of its children.

With this definition, we can specify a straightforward necessary and sufficient condition that ensures a DAG satisfies the CP conditions required by the definition of a UCP-net.

**Proposition 2** *Let $G$ be a DAG over $\{X_i\}$ whose factors $f_i$ reflect the GAI-structure of utility function $u$. Then $G$ is a UCP net iff each variable $X_i$ dominates its children.*

**Proof:** We need only show that the CP-condition holds for $X$ iff $X$ dominates its children. Assume the same variables stand in relation to $X$ as in the definition of domination above, and let $\mathbf{W} = \mathbf{V} - (\mathbf{U} \cup \{X\} \cup \mathbf{Y} \cup \mathbf{Z})$ (i.e., all of the other variables in the network). $X$ satisfies the CP-condition iff, for all $x_1, x_2, \mathbf{u}$: $x_1\mathbf{uyzw} \succeq x_2\mathbf{uyzw}$ implies $x_1\mathbf{u}(\mathbf{yzw})' \succeq x_2\mathbf{u}(\mathbf{yzw})'$ for all $(\mathbf{yzw})'$. Now $x_1\mathbf{uyzw} \succeq x_2\mathbf{uyzw}$ iff $u(x_1\mathbf{uyzw}) \geq u(x_2\mathbf{uyzw})$, iff

$$f_X(x_1, \mathbf{u}) - f_X(x_2, \mathbf{u}) \geq \sum_i f_{Y_i}(y_i, x_2, \mathbf{u}_i, \mathbf{z}_i) - f_{Y_i}(y_i, x_1, \mathbf{u}_i, \mathbf{z}_i)$$

since the only factors whose values can vary between these two terms are $f_X$ and the $f_{Y_i}$. By definition this relation holds for all values of $\mathbf{y}$ and $\mathbf{z}$ (and trivially for all $\mathbf{w}$) iff $X$ dominates its children. ◂

Determining if a quantified network is in fact a UCP-network requires a case-by-case analysis for each "extended family" in the network involving a number of tests exponential in the size of the extended families (by this we refer to a variable, its parents, its children, and its children's parents). Several stronger sufficient conditions exist that are easier to test. Here we present a particularly simple one.

**Proposition 3** *Let $G$ be a quantified DAG over the set of variables $V$. For each variable $X$ let $\mathbf{U}$ be its set of parents in $G$. For $x_1, x_2 \in Dom(X)$, let*

$$Minspan(x_1, x_2) = \min_{\mathbf{u} \in Dom(\mathbf{U})} (|f_X(x_1, \mathbf{u}) - f_X(x_2, \mathbf{u})|),$$
$$Minspan(X) = \min_{x_1, x_2 \in Dom(X)} Minspan(x_1, x_2).$$

*Define $Maxspan(X)$ analogously with $\max$ replacing $\min$. Then $G$ is a UCP-net if $Minspan(X) \geq \sum_i Maxspan(Y_i)$ where the $Y_i$ are the children of $X$.*

The values of *Maxspan* and *Minspan* can be computed for each variable $X$ in $O(|Dom(X)||f_X|)$ time. Thus, this condition can be checked in time polynomial in the number of network parameters.

For purposes of elicitation and computation, it is often convenient to normalize utilities over the range $[0, 1]$. Similarly, it is useful to normalize the individual factors of a UCP-net. In Section 5 we will consider a normalized variant of the UCP-net model in which the "rows" of each factor are normalized and "tradeoff" weights are used to calibrate them. Specifically:

(a) For each variable $X$, with parents $\mathbf{U}$ and factor $f_X$, and each $\mathbf{u} \in Dom(\mathbf{U})$, we normalize the function $f_X(x, \mathbf{u})$ so that its values lie in the range $[0, 1]$. That is, we insist that $\min_x f_X(x, \mathbf{u}) = 0$ and $\max_x f_X(x, \mathbf{u}) = 1$. We denote the normalized function $v_X^{\mathbf{u}}$ and call it the *local value function* for $X$ given $\mathbf{u}$.

(b) For each $X$ and instantiation of its parents $\mathbf{u}$, we specify a *multiplicative tradeoff weight* $\pi_X^{\mathbf{u}}$, and an *additive tradeoff weight* $\sigma_X^{\mathbf{u}}$.

---
[3]The example above can be accommodated by clustering the dependent variables. That is, we can define a new variable $C$ with $Dom(C) = Dom(A) \times Dom(B)$. In general, any utility function can be represented in a UCP-net by clustering appropriate sets of variables. This can be a reasonable approach if the clusters remain relatively small. It is also possible to generalize the definition of a UCP-net to allow cycles. This example could be represented by allowing $A$ and $B$ to be parents of one another. Cycles allow one to express a larger set of utility functions, but still do not permit all utility functions to be represented. They also may admit inconsistency: certain network structures with cycles do not correspond to a consistent utility function satisfying the CP-constraints (an impossibility in acyclic graphs). We refer to [3] for details.

The semantics of such a *normalized UCP-net* is as follows: the utility of any outcome is given by the sum of the terms (for each variable $X$) $\pi_X^{\mathbf{u}} v_X^{\mathbf{u}}(x) + \sigma_X^{\mathbf{u}}$, where $x$ is the instantiation of variable $X$ and $\mathbf{u}$ is the instantiation of its parents. It is not hard to see, by the usual transformation results in utility theory, that every UCP-net has an equivalent normalized representation.

## 4 Optimization Algorithms

The two types of queries discussed above, dominance queries and optimization queries, can both be answered directly in UCP nets. Dominance queries can be answered trivially: determining whether $u(\mathbf{v}_1) \geq u(\mathbf{v}_2)$ for two complete outcomes simply requires that one extract and sum the values of each factor in the network and then compare the sums. This can be done in time linear in the size of the network. Thus UCP-nets offer the advantages of other additive decompositions. In contrast, dominance testing in CP-nets is computationally difficult precisely because the tradeoffs between the (conditional) preference levels for different variables have not been specified.

Outcome optimization queries can also be answered directly given a UCP-net, taking linear time in the network size. Given a partial instantiation $\mathbf{z} \in Dom(\mathbf{Z})$, determining $\arg\max\{u(o) : o \in Comp(\mathbf{z})\}$ can be effected by a straightforward sweep through the network. Let $X_1, \ldots, X_n$ be any topological ordering of the network variables excluding $\mathbf{Z}$. We set $\mathbf{Z} = \mathbf{z}$, and instantiate each $X_i$ in turn to its maximal value given the instantiation of its parents. This procedure exploits the considerable power of the CP semantics to easily find an optimal outcome given certain observed evidence (or imposed constraints).

**Proposition 4** *The forward sweep procedure constructs the optimal outcome $\arg\max\{u(\mathbf{v}) : \mathbf{v} \in Comp(\mathbf{z})\}$.*

**Proof:** Let $\mathbf{v}_\mathbf{z}$ be any outcome in the set of completions of $\mathbf{z}$. Define a sequence of outcomes $\mathbf{v}_i, 0 \leq i \leq n$, as follows: (a) $\mathbf{v}_0 = \mathbf{v}_\mathbf{z}$; (b) if $X_i \notin \mathbf{Z}$, $\mathbf{v}_i$ is constructed by setting the value of $X_i$ to its most preferred value given the instantiation of its parents in $\mathbf{v}_{i-1}$, with all other variables retaining their values from $\mathbf{v}_{i-1}$; (c) if $X_i \in \mathbf{Z}$, then $\mathbf{v}_i = \mathbf{v}_{i-1}$. By construction, $\mathbf{v}_i \succeq \mathbf{v}_{i-1}$. The last outcome $\mathbf{v}_n$ is precisely that constructed by the forward sweep algorithm (assuming ties are broken in the same way as in the procedure). Since $\mathbf{v}_n \succeq \mathbf{v}_\mathbf{z}$ for any outcome $\mathbf{v}_\mathbf{z}$ consistent with the evidence, the forward sweep procedure yields an optimal outcome. ◄

This algorithm illustrates the sharp contrast between UCP-nets and GAI representations. Effective outcome optimization in a GAI model requires that one use a dynamic programming algorithm like variable elimination. As a consequence, the complexity of such an algorithm—exponential in the induced tree width of the GAI model—depends critically on the "topology" of the model and the ability to find good elimination orderings.

Thus, UCP-nets offer a valuable restriction of GAI models and generalization of CP-nets. In particular, they impose restrictions on the relative strength of the GAI factors, and generalize CP-nets to allow for the representation of quantitative utilities. But we preserve the convenient graphical representation of CP-nets, and gain considerable computational benefits over both models.

One of the main reasons to move from qualitative to quantitative preference models is to support decisions under (quantified) uncertainty. Naturally, given a decision problem and a fully specified UCP-network, determining an optimal course of action is (conceptually) straightforward. When the distributions induced by actions can be structured in a Bayes net, UCP-nets can be used to help structure the decision problem. Suppose that the distribution over variables $\mathbf{V}$ determined by an action $a$ is represented as a Bayes net (possibly with a choice node if we wish to represent all actions) and the utility function over outcome space is determined by a UCP-net.[4] To compute the optimal action, we can construct an influence diagram by adding one *utility variable* $F_i$ for each (nonconstant) factor $f_{X_i}$ in the UCP-network. $F_i$ has as parents both $X_i$ and the parents of $X_i$ in the UCP-net, and is quantified using factor $f_{X_i}$ from the UCP-net. Variable elimination (e.g., Dechter's [6] MEU variant) can be used to determine the optimal action.[5]

This approach uses the GAI factorization of utilities afforded by the UCP-net, but not the CP-semantics. We can improve upon these ideas by noticing that our goal is to select the optimal action, not (necessarily) compute its expected utility. In any GAI representation, we can bound the error associated with ignoring a utility variable $F_i$ with parents $\mathbf{U}$ as follows. Let $\varepsilon_i = \max_{\mathbf{u}} F_i(\mathbf{u}) - \min_{\mathbf{u}} F_i(\mathbf{u})$. The expected value $EV(a)$ of any action $a$ is given by

$$\sum_{\mathbf{v} \in \mathbf{V}} Pr(\mathbf{v}|a)(\sum_i F_i(\mathbf{v})) = \sum_i (\sum_{\mathbf{v} \in \mathbf{V}} Pr(\mathbf{v}|a) F_i(\mathbf{v}))$$

---

[4] Often utilities are elicited only over variables that are directly related to preferences. The variables in the Bayes net may include variables not contained in the UCP-net.

[5] One might also consider how expected utility networks [11] might be used in this regard.

Let $EV_{-i}(a)$ be the expected value of action $a$ with respect to all utility variables except $F_i$. Then $|EV(a) - EV_{-i}(a)| \leq \varepsilon_i$. Thus if there is some $a^*$ such that $EV_{-i}(a^*) - EV_{-i}(a) \geq \varepsilon_i$ for all $a \neq a^*$, we know $a^*$ is optimal without having to compute the $i$'th term in the above summation. Analogous statements hold for ignoring any subset $\mathbf{F}$ of the utility variables, setting $\varepsilon_{\mathbf{F}} = \sum_{i \in \mathbf{F}} \varepsilon_i$.

This suggests an incremental technique for computing an optimal (or near-optimal) action that exploits the CP-semantics of a UCP-net. Let $X_1, \ldots X_n$ be a topological ordering of the variables in the UCP-net. Our technique runs in (at most) $n$ stages, where at stage $k$, we compute $EV_{-\{i>k\}}(a)$ for each action $a$. If for some $a^*$ we have $EV_{-\{i>k\}}(a^*) - EV_{-\{i>k\}}(a) \geq \varepsilon_{\{i>k\}}$ for all $a \neq a^*$, we know $a^*$ is optimal and we can terminate without computing any further terms. Furthermore, we can remove from consideration at subsequent stages any action whose partial utility differs by more than $\varepsilon_{\{i>k\}}$ from the (estimated) optimal action at this stage. The motivation for this approach lies in the fact that in a UCP network, variables near the top of the UCP-net have a larger impact on utility, and are thus more likely to lead to the separation of actions than factors lower in the net. For example, if action $a$ has high probability of making the most important variable $X_1$ take its most desired value, while action $b$ is likely to ensure its least desired value, we may be able to eliminate $b$ from consideration by just computing the first term of above summation. We can also terminate when the error associated with $a^*$ is below some threshold, even if it is not optimal.

The computational benefits arise when one considers that computing $EV_{-\{i\geq k\}}(a)$ requires one to do inference only on those variables that are relevant to predicting $\mathbf{F}_{-\{i>k\}}$. Furthermore, at each each stage we need only compute the expected value with respect to the newly added utility variable. In a problem with no evidence, for example, this can be accomplished by considering only ancestors of the variables $F_i$ in the Bayes net. An important issue with this iterative procedure is how to minimize overall computation by reusing information computed in earlier iterations. This is plausible since (utility) factors generally overlap, and these factors generally have common influences. For example, if VE is used it might be possible to find variable orderings for each computation that facilitate the reuse of previous results—determining such orderings is an interesting algorithmic question.

## 5　Elicitation with UCP-nets

One of the key problems facing the use of decision-theoretic models is the elicitation of preference information. In this section we describe one possible procedure for exploiting the structure and semantics of a UCP-network to facilitate an incremental elicitation process. More precisely, given a specific decision scenario—i.e., a set of possible actions and the corresponding distributions over outcomes they induce—and a set of constraints on the tradeoff weights of a normalized UCP-network, the *regret* of the best action can be computed as a simple linear optimization problem. Given a specific set of questions that can be posed to the user, the (myopic) value of information for each question with respect its reduction of the minimax regret can also be computed by solving a linear program. As such, an incremental procedure can be used to compute a greedy query plan that will ask just enough questions of the user to decide on a course of action whose regret is below some prespecified threshold (if this threshold is set to zero, then the optimal action will be recommended).

To make this more precise, we define a *decision scenario* to be a set of actions $\mathcal{A} = a_1, \ldots, a_n$, where each action $a_i \in \mathcal{A}$ induces a distribution $Pr_i(\mathbf{V})$ over outcomes. Let $\mathcal{O}_i$ denote the support set for $Pr_i$ (i.e., the set of outcomes $\mathbf{v}$ for which $Pr_i(\mathbf{V}) > 0$). We generally assume that $\mathcal{O}_i$ is small relative to $Dom(\mathbf{V})$.

As a starting point, we assume that we have been provided with a normalized UCP-network, whose structure and local value functions $v_X^{\mathbf{u}}$ have been provided, but whose tradeoff weights $\pi_X^{\mathbf{u}}$ and $\sigma_X$ remain unspecified. We feel that the elicitation of both structure and local value functions is something that users will often be able to provide without too much difficulty. Structure elicitation, involving questions regarding the relative importance of different attributes, as well as the dependence of these assessments on other attributes should not be especially onerous. Eliciting value function $v_X^{\mathbf{u}}$ requires ordering a small number of values, given a specific parent context, and calibrating these value using (local) standard gambles, again, a relatively unproblematic task.

Although the structure and value functions are thus determined, the tradeoff weights for the normalized UCP-network remain unknown. The utility of any outcome $\mathbf{v}$ is a linear function of these weights: specifically, if $\mathbf{v}$ instantiates each variable $X_i$ to $x_i$ and parent set $\mathbf{U}_i$ to $\mathbf{u}_i$, then

$$u(\mathbf{v}) = \sum_i \pi_{X_i}^{\mathbf{u}_i} v_{X_i}^{\mathbf{u}_i}(x_i) + \sigma_{X_i}^{\mathbf{u}_i}$$

Let $\mathbf{W}$ be the set of possible instantiations of the tradeoff weights, and $u(\mathbf{v}, \mathbf{w})$ denote the utility of $\mathbf{v}$ for a particular instantiation $\mathbf{w} \in \mathbf{W}$. The expected value of any action $a_i$ is also a linear function of the tradeoff weights:

$$EV(a_i, \mathbf{w}) = \sum_{\mathbf{v} \in \mathcal{O}_i} Pr_i(\mathbf{v}) u(\mathbf{v}, \mathbf{w})$$

Note that by assumption $|\mathcal{O}_i|$ is relatively small, so this sum should contain only a small number of terms.

We define the optimal action $a_{\mathbf{w}}^*$ with respect to an instantiation $\mathbf{w}$ of the tradeoff weights to be

$$a_{\mathbf{w}}^* = \arg\max_{a_i} EV(a_i, \mathbf{w})$$

If the utility function were known to have weights $\mathbf{w}$, $a_{\mathbf{w}}^*$ would be the correct choice of action. The *regret* of action $a_i$ with respect to $\mathbf{w}$ is

$$R(a_i, \mathbf{w}) = EV(a_{\mathbf{w}}^*, \mathbf{w}) - EV(a_i, \mathbf{w})$$

i.e., the loss associated with executing $a_i$ instead of acting optimally. Let $C$ be a subset of the set of possible instantiations of the tradeoff weights, $\mathbf{W}$. We define the *maximum regret* of action $a_i$ with respect to $C$ to be

$$MR(a_i, C) = \max_{\mathbf{w} \in C} R(a_i, \mathbf{w})$$

Finally, we define the action $a_C^*$ with *minimax regret* with respect to $C$:

$$a_C^* = \arg\min_{a_i} MR(a_i, C)$$

The *(minimax) regret level* of weight set $C$ is $MMR(C) = MR(a_C^*, C)$. If the only information we have about a user's utility function is that it lies in the set $C$, then $a_C^*$ is a reasonable choice of action. Specifically, without distributional information over the set of possible utility functions, choosing (or recommending) $a_C^*$ minimizes the worst case loss with respect to possible realizations of the utility function.

If $C$ is defined by a set of linear constraints on the weights, then $a_C^*$ as well as $MMR(C)$ can be computed using a set of linear programs. First note that we can compute

$$\max_{\mathbf{w} \in C} EV(a_j, \mathbf{w}) - EV(a_i, \mathbf{w})$$

for any pair of actions $a_i$ and $a_j$ using a linear program: we are maximizing a linear function of the weights subject to the linear constraints that define $C$. Solving $O(n^2)$ such linear programs, one for each ordered pair of actions, allows us to identify the action $a_C^*$ that achieves minimax regret and to determine the minimax regret level $MMR(C)$.

$MMR(C)$ tells us how bad off we could be recommending $a_C^*$. In particular if $MMR(C) = MR(a_C^*, C) = 0$ then $a_C^*$ dominates all other actions: it is at least as good as any other action for every feasible set of tradeoff weights. However, unless $C$ is very refined (i.e., is defined by strong constraints), multiple actions will potentially be optimal (i.e., will be maximal in certain regions of weight space). To determine which of these actions to recommend, we need to refine the constraints defining $C$ further.

$C$ can contain a range of different linear constraints. One class of constraints in $C$ is imposed by the structure of the network. In particular, each variable must dominate its children: this is a necessary condition in any UCP-net. Using the same notation for variable $X$'s neighborhood as in Defn. 2, domination imposes the following linear constraint on the weights: for each $\mathbf{u}, \mathbf{z}, \mathbf{y}$ and pair $x_1, x_2 \in Dom(X)$ such that $v_X^\mathbf{u}(x_1) \geq v_X^\mathbf{u}(x_2)$, we must have:

$$\pi_X^\mathbf{u} v_X^\mathbf{u}(x_1) - \pi_X^\mathbf{u} v_X^\mathbf{u}(x_2)$$
$$\geq \sum_{Y_i} \left( \begin{array}{c} \pi_{Y_i}^{x_2 \mathbf{u}_i \mathbf{z}_i} v_{Y_i}^{x_2 \mathbf{u}_i \mathbf{z}_i}(y_i) + \sigma_{Y_i}^{x_2 \mathbf{u}_i \mathbf{z}_i} \\ - \pi_{Y_i}^{x_1 \mathbf{u}_i \mathbf{z}_i} v_{Y_i}^{x_1 \mathbf{u}_i \mathbf{z}_i}(y_i) - \sigma_{Y_i}^{x_1 \mathbf{u}_i \mathbf{z}_i} \end{array} \right)$$

Another class of constraints is the set of bounds that restrict each tradeoff weight $\pi_X^\mathbf{u}$ and $\sigma_X^\mathbf{u}$ to a specific range. Such bounds are required in order to keep the LP problems we have proposed using bounded. Eliciting such bounds from the user is not a difficult task, as one can always start with very loose bounds. For example, the minimum and maximum utility of any possible outcome is a simple uniform bound on the tradeoff weights. Besides these required constraints, $C$ could contain other *nonstructural* linear constraints provided by the user, e.g., constraints on the relative magnitudes of different weights (reflecting degree of importance) or constraints on the relative expected utility of different actions in certain fixed contexts.

If minimax regret is zero, or lies below some acceptable threshold, the action $a_C^*$ can be recommended. Otherwise, questions can be asked of the user to help differentiate between possible actions. The solution to the above set of $O(n^2)$ linear programs can provide some guidance. For example, the linear program for solving $MR(a_i, C)$ also yields a solution to the dual problem. This solution provides a multiplicative factor associated with every inequality in $C$ that tells us how much of a change we can produce in $MR(a_i, C)$ (in the neighborhood of the optimal solution) by modifying the inequality. Say that the $k$-th inequality is the upper bound $\pi_X^U \leq 290$, and that the value of the $k$-th variable in the dual solution is 100. This tells us that if we can get the user to tighten their upper bound on $\pi_X^U$ by 9 units this might yield a 900 unit decrease in $MR(a_i, C)$. By examining the factors associated with the bounds imposed on the weights, those weights that have the most potential to influence $MMR(C)$ could be identified. Furthermore, after we have queried the user and obtained an updated bound we need not resolve the linear programs from scratch to recompute the maximum regret of each action. There are many techniques in the LP literature on sensitivity analysis that can be employed to minimize the amount of computation that needs to be performed [5].

However, generally for realistic elicitation we cannot rely solely on the recommendations of the linear programs. In particular, sharpening the inequality recommended by the dual solution might not be an easy task for the user. The types of questions that can (reasonably) be asked will be domain dependent, and influenced by factors such as the complexity of the domain (e.g., if the number of attributes is manageable, asking a user to compare full outcomes may be acceptable, but otherwise not), the sophistication of the user, and the importance of the decision to be made. To address the general problem here, we will assume a (finite) set of possible questions $Q = \{q_1, \ldots, q_k\}$, with each $q_i$ having $m$ possible responses $r_i^1, \ldots, r_i^m$ (we fix the number of responses simply to streamline the presentation). We suppose that every response adds an additional linear constraint to $C$ (this subsumes the case of sharpening an existing constraint). Let $C(r_i^j)$ denote the set of weights that satisfy $C \cup \{r_i^j\}$. Then asking a question $q_i$ and receiving a response $r_i^j$ will reduce minimax regret by the amount $MMR(C) - MMR(C(r_i^j))$.

This suggests a querying strategy in which questions that have the ability to reduce minimax regret the most are asked first. In a certain sense, asking questions that reduce minimax regret can be seen as a distribution-free analog of traditional value of information approaches to querying. Specifically, the procedure we suggest strongly parallels the elicitation method proposed in [4], where a distribution over possible utility functions is used to guide the interactive elicitation process. In our distribution-free model, we cannot define the expected value of a question, but instead use the worst-case response to define the minimal improvement we can obtain from some question. The *minimax improvement* of question $q_i$, $MI(q_i)$ is $\min_j MMR(C) - MMR(C(r_i^j))$. The *minimax optimal query* with respect to $C$ is that query with maximal improvement $MI(q_i)$.[6] We note that the improvement $MI(q_i)$ for any query must be nonnegative, since $q_i$ will always reduce the size of the feasible weight space, and generally will be nonzero.

This suggests the following abstract elicitation strategy, which myopically attempts to improve minimax regret. Given a set of feasible weights $C$, the query $q_i$ with maximal improvement is asked, and response $r_i^j$ is obtained, resulting in a more refined weight space $C(r_i^j)$. Then $MMR(C(r_i^j))$ is computed by solving the previously specified linear programs with the added constraint $r_i^j$. Techniques for sensitivity analysis can be utilized to minimize the work involved in doing these computation. This process is repeated until one of two conditions is met: (a) the current weight space admits an action with regret less than some threshold $\tau$; or (b) no query has an improvement score greater than the cost of that query. This latter condition is typically important in interactive elicitation: while one could ask many questions to narrow down a utility function so that a (near) optimal decision can be made, one must account for the cost of these questions (e.g., the burden they impose on the user).

Making this procedure concrete requires having a set of possible questions whose responses induce linear constraints on weight space. As pointed out above, such questions will in general be domain dependent. However, they might include asking the user to quantify the relative strengths of various tradeoff weights associated with a single variable $\sigma_X$ and $\pi_X$. For example, asking the user for a value of $k$ such that $\sigma_X^{u_1} \leq k\sigma_X^{u_2}$. Since this involves the outcomes of a common variable it should be relatively easy for the user to answer. Assessing relative tradeoff weights associated with different variables is a similar, albeit more difficult, question. Sharpening an upper or lower bound on a weight was addressed above. One more type of question might be to ask the user which of actions $a_i$ or $a_j$ she would prefer in a *specific* context. The answer to this question again imposes a linear constraint on weight space.

This approach is very similar to that utilized in *imprecisely specified multiattribute utility theory (ISMAUT)* [9]. In such work, standard additive independent utility models are assumed and constraints on tradeoffs weights are used to determine if an optimal decision can be made. If no action dominates all others, further preferences are elicited. Our approach extends this basic viewpoint in a number of ways, including the utilization of richer utility models and a minimax regret model that supports decisions even when no action is dominant. ISMAUT does not generally describe means for generating queries automatically or making decisions when no action is dominant.

Our elicitation procedure has several drawbacks. The greedy nature of the algorithm means that *sequences* of questions that reduce regret may not be considered if the individual questions in the sequence do not. Circumventing this requires lookahead or some form of dynamic programming. This problem is common to most value of information approaches (e.g., [4]). The use of minimax regret to select actions should be viewed as reasonable in the absence of distributional information. However, selecting queries so that the *worst* response has maximal improvement may not always be appropriate; and comparing this worst case improvement to the cost of the query may also be questioned, but other strategies are possible.

We are currently exploring the use of distributions over weight space (i.e., over utility functions) to guide the elicitation process. In the abstract, such a model would be similar to that of Chajewska, Koller, and Parr [4]. The differences lie in the use of UCP-nets as the underlying utility representation, and the use of dynamic programming to construct optimal query sequences (rather than using myopic value of information). Integrating this with linear optimization poses some interesting challenges.

## 6　Concluding Remarks

We have proposed a new directed graphical model for representing utility functions that combines appealing aspects of both CP-nets and GAI models. The UCP-net formalism has a number of conceptual and computational advantages over these models, providing leverage with respect to inference and elicitation. Clearly, practical experience and empirical studies are needed to gauge the ultimate effectiveness of UCP-nets. However, we are encouraged by the widespread use of additive models, and more recently, by the successful application of CP-nets to the problem of preference-based Web page content configuration [8].

We are currently in the process of implementing the interactive elicitation algorithm described in Section 5. Future research includes the inclusion of distributional information over utility functions (or tradeoff weights), and developing algorithms that compute and use value of information to construct optimal query plans. Another fundamental question pertains to determining optimal query plans when the query space is large or infinite, involving parameterized queries (e.g., standard gambles). We expect that the considerable structure exhibited by the problem (e.g., the fact that the set of actions divides weight space $\mathbf{W}$ into a set of convex regions where each action is optimal) will allow optimization over each query *type* to be effected, without explicit enumeration of all instances.

---

[6] Queries can be ranked simply using their worst-case minimax regret, rather than their worst-case improvement, since the term $MMR(C)$ is common to all queries and responses.

The investigation of the suggested optimization algorithms, specifically, empirical validation of the incremental variable elimination approach to decision making described in Section 4, is a high priority.

## Acknowledgements

Craig Boutilier and Fahiem Bacchus were supported by Communications and Information Technology Ontario, the Institute for Robotics and Intelligent Systems, and the Natural Sciences and Engineering Research Council of Canada. Ronen Brafman was supported by the Paul Ivanier Center for Robotics and Production Management.